\documentclass[12pt]{article}

\usepackage[dvips]{graphicx}
\usepackage{natbib}
\usepackage[latin1]{inputenc}
\usepackage{textpos}

\title{Perspective Alignment in Spatial Language}
\author{Luc Steels,$^{1,2\ast}$ Martin Loetzsch$^{1}$\\
\\
\normalsize{$^{1}$Sony Computer Science Laboratory - Paris}\\
\normalsize{6, rue Amyot, 75005 Paris - France}\\
\normalsize{$^{2}$VUB AI Lab, Vrije Universiteit Brussels}\\
\normalsize{Pleinlaan 2, 1050 Brussels - Belgium}\\
\\
\normalsize{$^\ast$E-mail: steels@arti.vub.ac.be}
}

\date{}

\begin{document}
\DeclareGraphicsExtensions{.jpg,.ps,.eps}

\maketitle 

\begin{textblock}{9}(0,-6)
{\tt\footnotesize\noindent To appear in: K. R. Coventry, T. Tenbrink, and J. A. Bateman, editors, Spatial Language and Dialogue. Oxford University Press, 2008}
\end{textblock} 

\begin{abstract}
  It is well known that perspective alignment plays a major role in
  the planning and interpretation of spatial language. In order to
  understand the role of perspective alignment and the cognitive
  processes involved, we have made precise complete cognitive models
  of situated embodied agents that self-organise a communication
  system for dialoging about the position and movement of real world
  objects in their immediate surroundings.  We show in a series of
  robotic experiments which cognitive mechanisms are necessary and
  sufficient to achieve successful spatial language and why and how
  perspective alignment can take place, either implicitly or based on
  explicit marking.
\end{abstract}

\section{Introduction}

Spatial language consists of expressions that involve spatial positions and movements of objects in the world. Spatial language always involves perspective \citep{schober93spatial,schober08spatial}. For example, the meaning of the phrase ``the ball left of the glass" depends on the spatial position of the viewer with respect to the objects involved. Moreover this viewer can be the speaker (egocentric) or the hearer or somebody else involved in the conversation (allocentric). In any case, if speaker and hearer perceive a scene from different perspectives, they need to align the perspective from which the scene is being described in order to make sense of the description. Often perspective is implicit and dialogue partners must then indirectly align perspective. But natural languages have also various ways to make perspective explicit, as in ``the ball to my left" (see also \citeauthor{carlson08formulating}, this volume). 

The goal of our work is to explain these well known facts. Concretely, we would like to understand why perspective is unavoidable in spatial language, how dialogue partners can still align perspective even if it is not marked, and why and how marking helps. We would also like to understand how the whole system can come off the ground, in other words how spatial language involving implicit or explicit perspective alignment can be learned or invented through negotiation in consecutive dialogues. 

Our explanations will be based on making very precise and complete models of communicating embodied agents, situated in a particular real world environment. The models are complete in the sense that they include mechanisms for achieving physical behavior in the real world, vision for the construction of situation models, cognitive mechanisms for developing and using spatial categories like left/right, forward/backward, close/far, and mechanisms for developing and using  lexicons. Our models have been completely formalised and implemented on physical robots so that we can test their effectiveness and behavior in repeatable experiments. In each experiment, we set the agents up to play situated language games in the form of dialogues about the objects in their world. The agents describe to each other the movement of a ball in their close proximity. Because spatial language is obviously a very useful and effective way to do so, we expect it to emerge as part of consecutive games. 

We will make three arguments: 

\begin{enumerate}
\item {\it As soon as agents are embodied, they necessarily have a specific view on the world and  spatial language becomes impossible without considering perspective.} We will show this by an experiment in which first the agents see the world through the same camera (in other words two agents use the same robot body) and hence they have exactly the same visually derived situation model. And second the agents are made to see the world through their own camera and so they each have a different situation model. The experiment clearly shows that in the second case, a communication system cannot come off the ground: They cannot learn the meaning of spatial terms from each other, and they generally fail to understand each other. 

\item {\it Perspective alignment is possible when the agents are endowed with two abilities: (i) to see where the other one is located, and (ii) to perform a geometric transformation known as Egocentric Perspective Transform.} This transformation allows the agent to compute what the scene looks like from the viewpoint of the other, in other words to develop a situation model from the other partner's perspective. The Egocentric Perspective Transform is normally carried out in the parietal-temporal-occipital junction \citep{zacks99imagined} and used for a wide variety of non-linguistic
tasks, such as prediction of the behavior of others or navigation \citep{iachini03role}. We have implemented these capabilities and performed an experiment in which agents test systematically from which perspective an utterance makes sense. They are thus able to implicitly align perspective, but only because they are both grounded and situated in the same real world setting. The experiment demonstrates that agents are in this case able to bootstrap spatial language and achieve successful communication. Note that this is still without explicitly marking perspective. 

\item {\it Perspective alignment takes less cognitive effort if perspective is marked.} We investigate this through another experiment that compares the implicit way of perspective alignment (as in (2)) with one where perspective becomes marked because the lexical processes now express to what perspective the speaker/hearer is aligned (egocentric or allocentric).  We observe a significant decrease of cognitive effort. This experiment shows additionally that our models are adequate for demonstrating how perspective markers can be invented and learned. This is not a simple problem and children can only do it fairly late in language development. 

\end{enumerate}

The remainder of the paper is in two parts. The first part gives more details on the experimental setup and on the various cognitive mechanisms that make up the agent architecture. The second part reports results of our experiments. A final part of the paper derives some conclusions. 

\section{Experimental Setup}

\begin{figure}[t]
\centerline{\includegraphics[width=0.8\columnwidth]{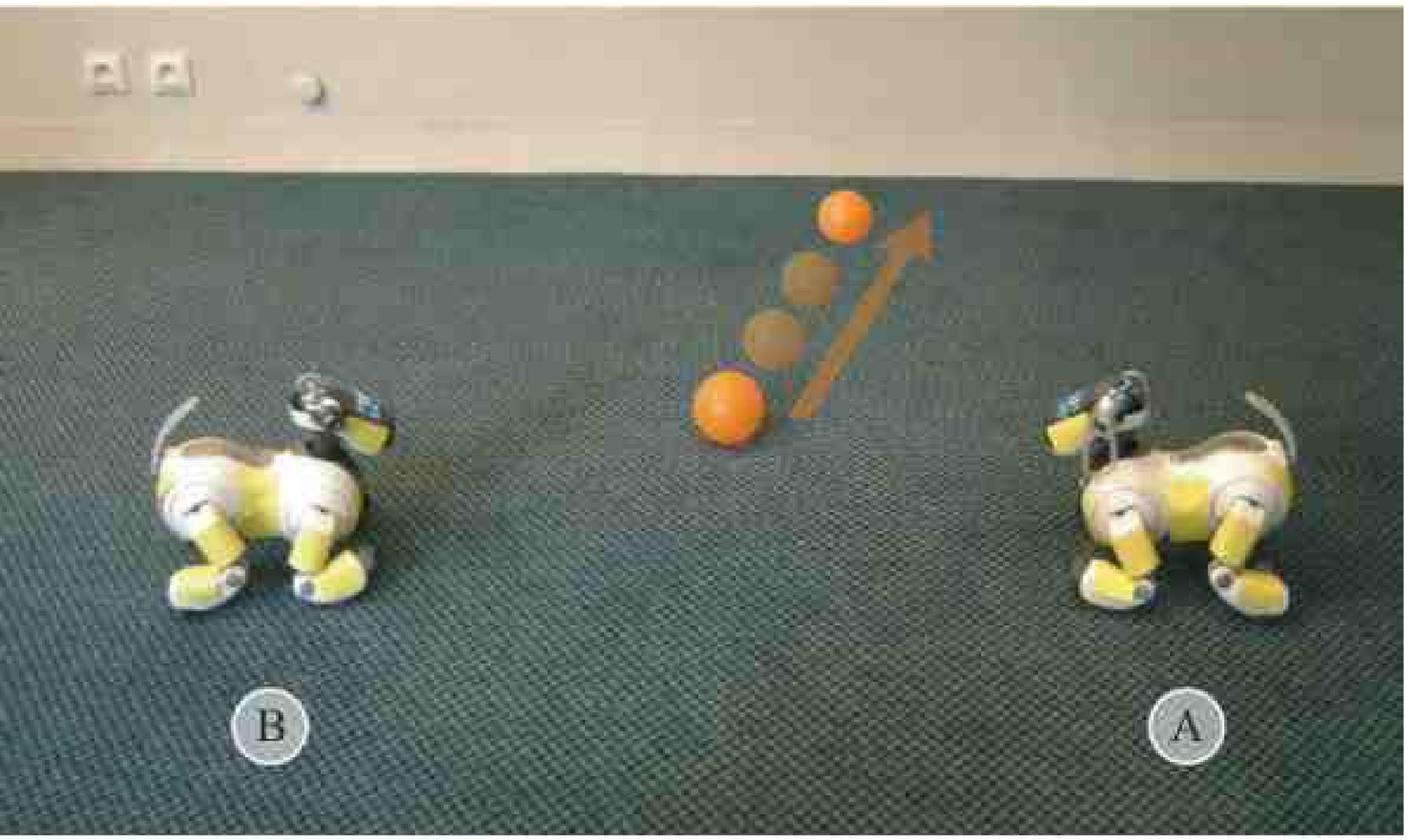}}
\caption{Agents embodied in physical robots. The speaker (in robot $A$) and the hearer (in robot $B$) together observe ball movement events and then play a language game to describe the scene to each other.}
\label{f:scene}
\end{figure}

A lot of work has recently been done on studying human dialogue \citep{Clark:1996,PickeringGarrod:2004}. The methodological approach discussed here is entirely complementary. We take the findings of these investigations as given but try to see what it takes to build synthetic models of dialogue, which obviously requires a `mechanistic' theory of all the processes involved in dialogue and a concrete setup where we can test these processes. Moreover we are interested to understand how spatial language with perspective marking can arise in a population, motivated by attempts to understand the origins and evolution of communication systems \citep{steels03evolving}. 

\begin{figure}[p]
\centerline{
\begin{tabular}{ll}
a) & b) \\
\includegraphics[width=0.4\columnwidth]{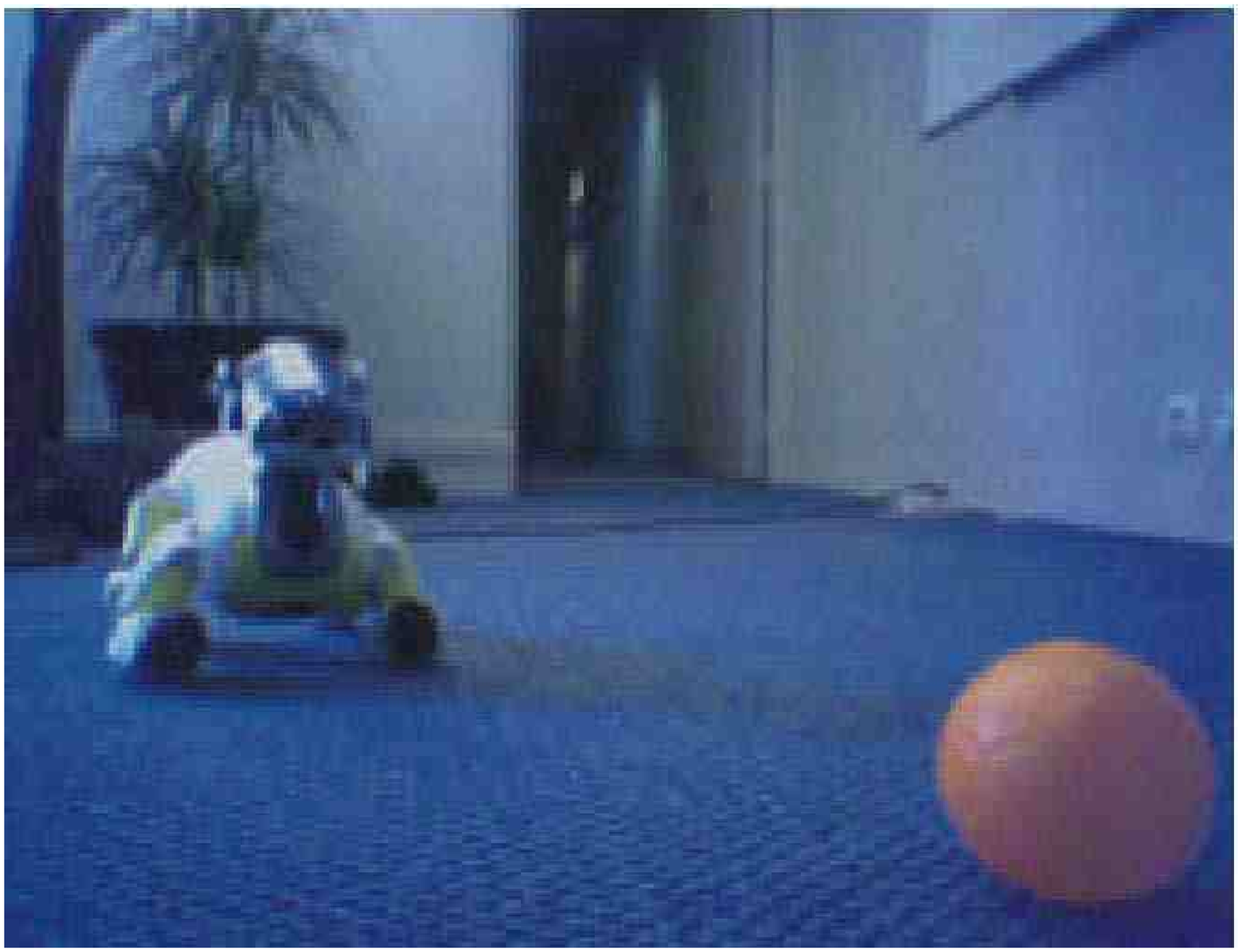} &
\includegraphics[width=0.4\columnwidth]{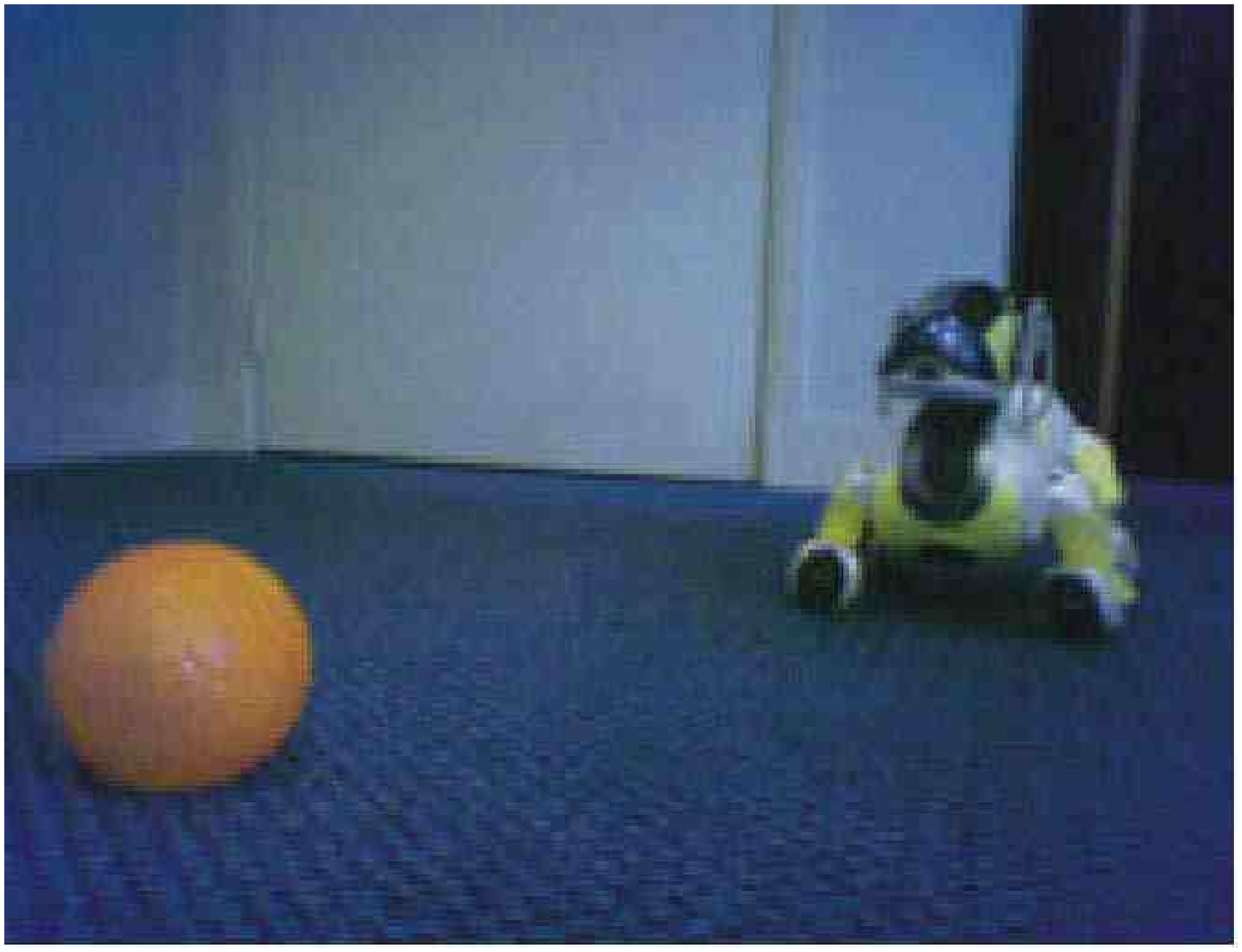} \\
c) & d) \\
\includegraphics[width=0.4\columnwidth]{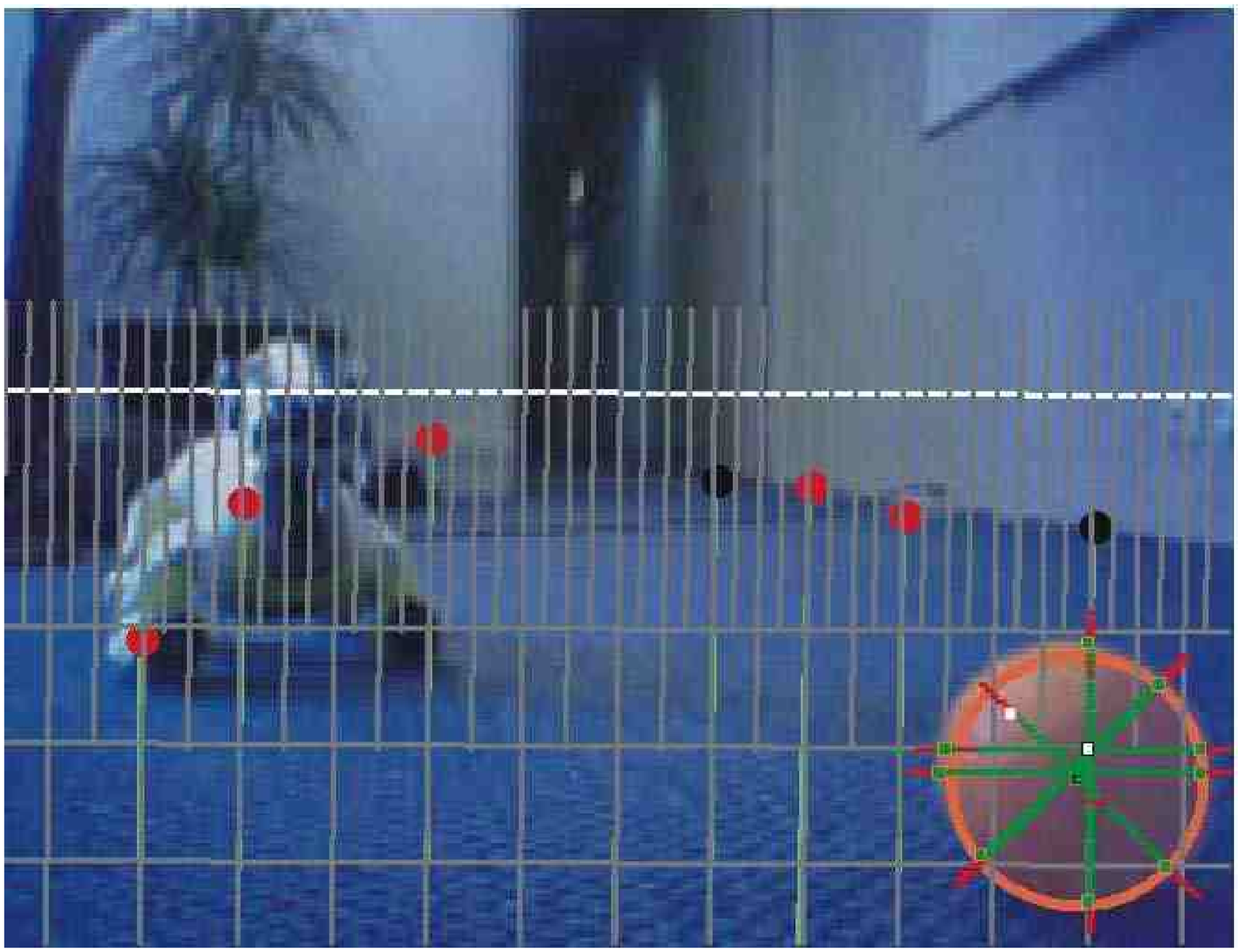} &
\includegraphics[width=0.4\columnwidth]{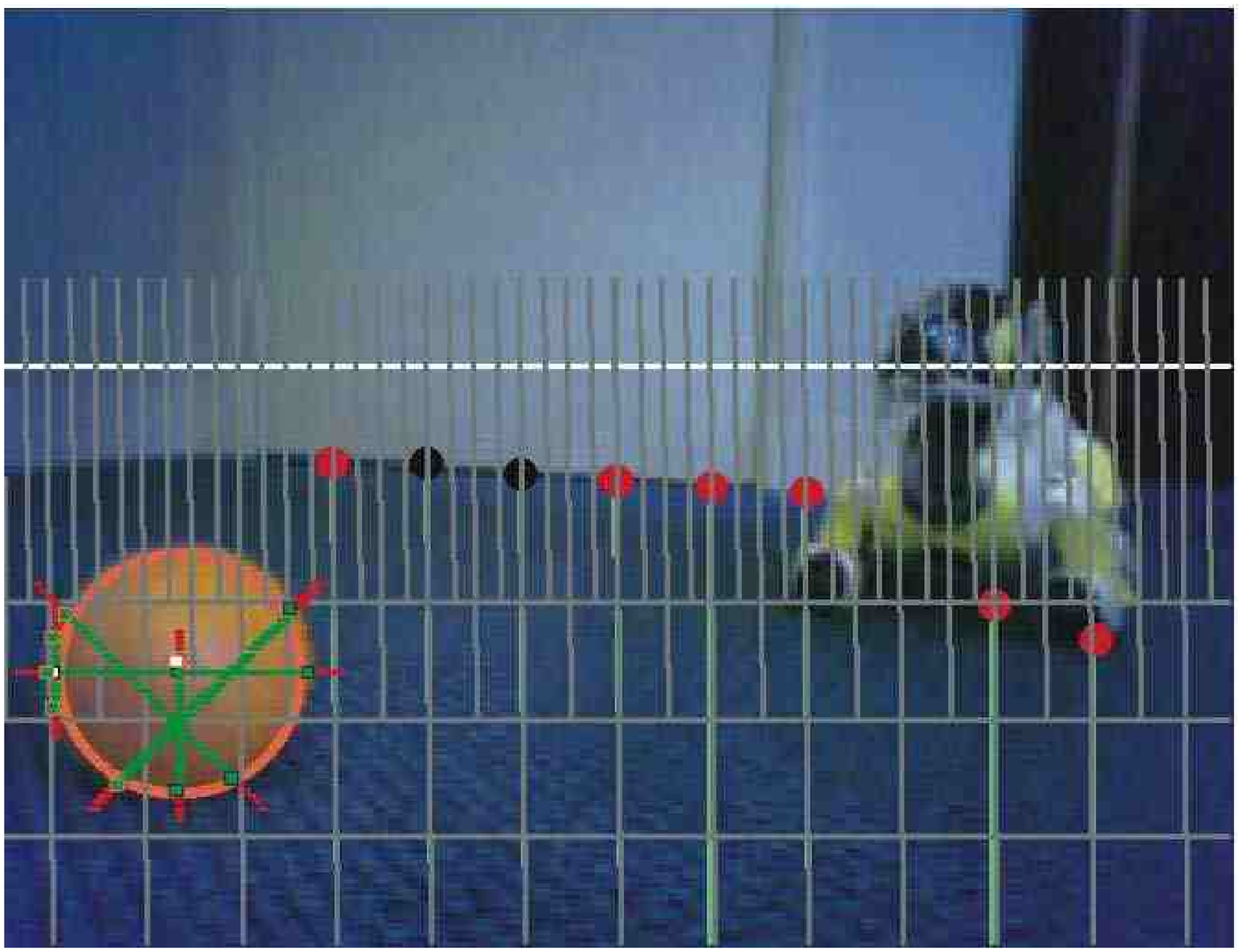} \\
e) & f) \\
\includegraphics[width=0.4\columnwidth]{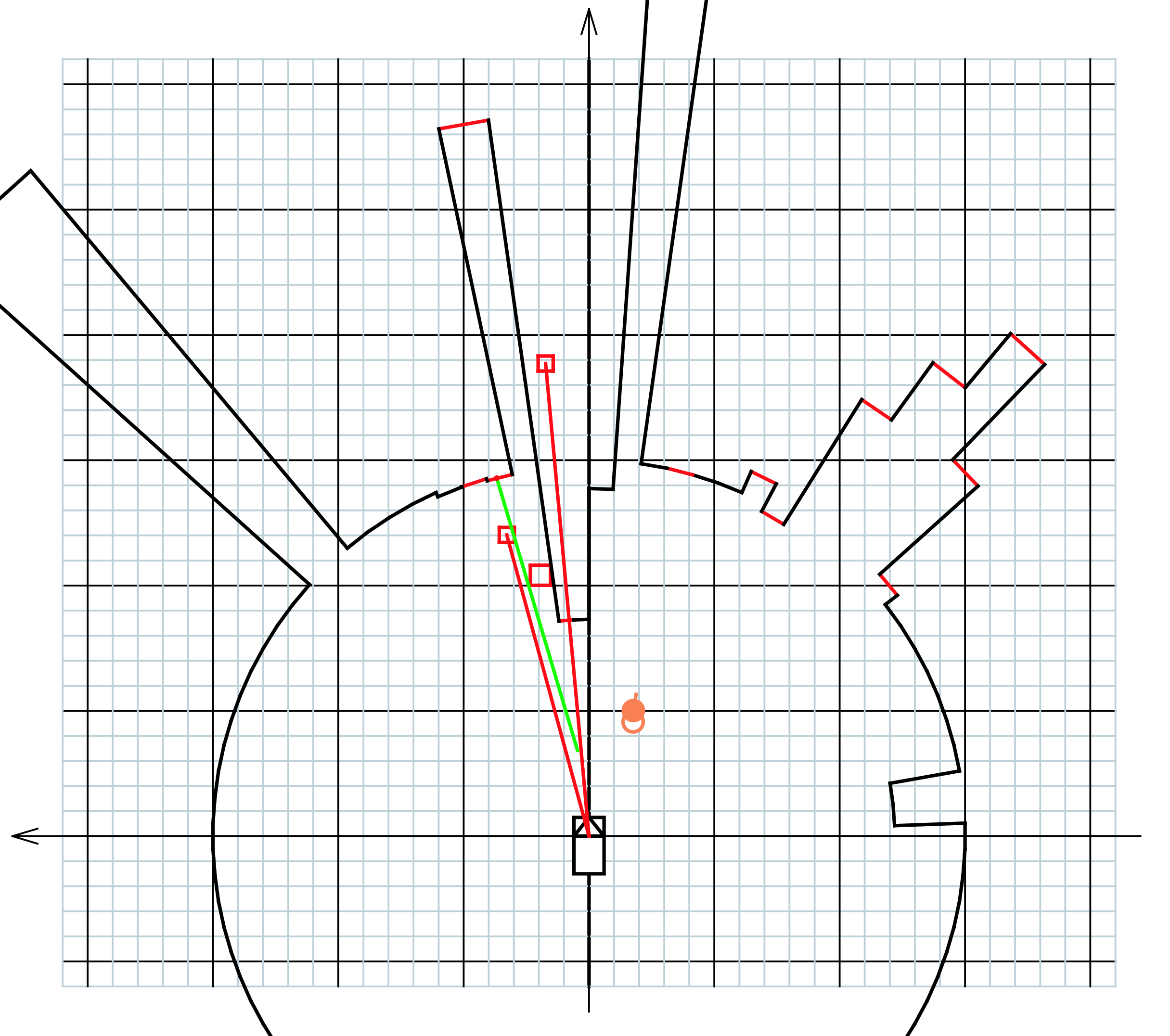} &
\includegraphics[width=0.4\columnwidth]{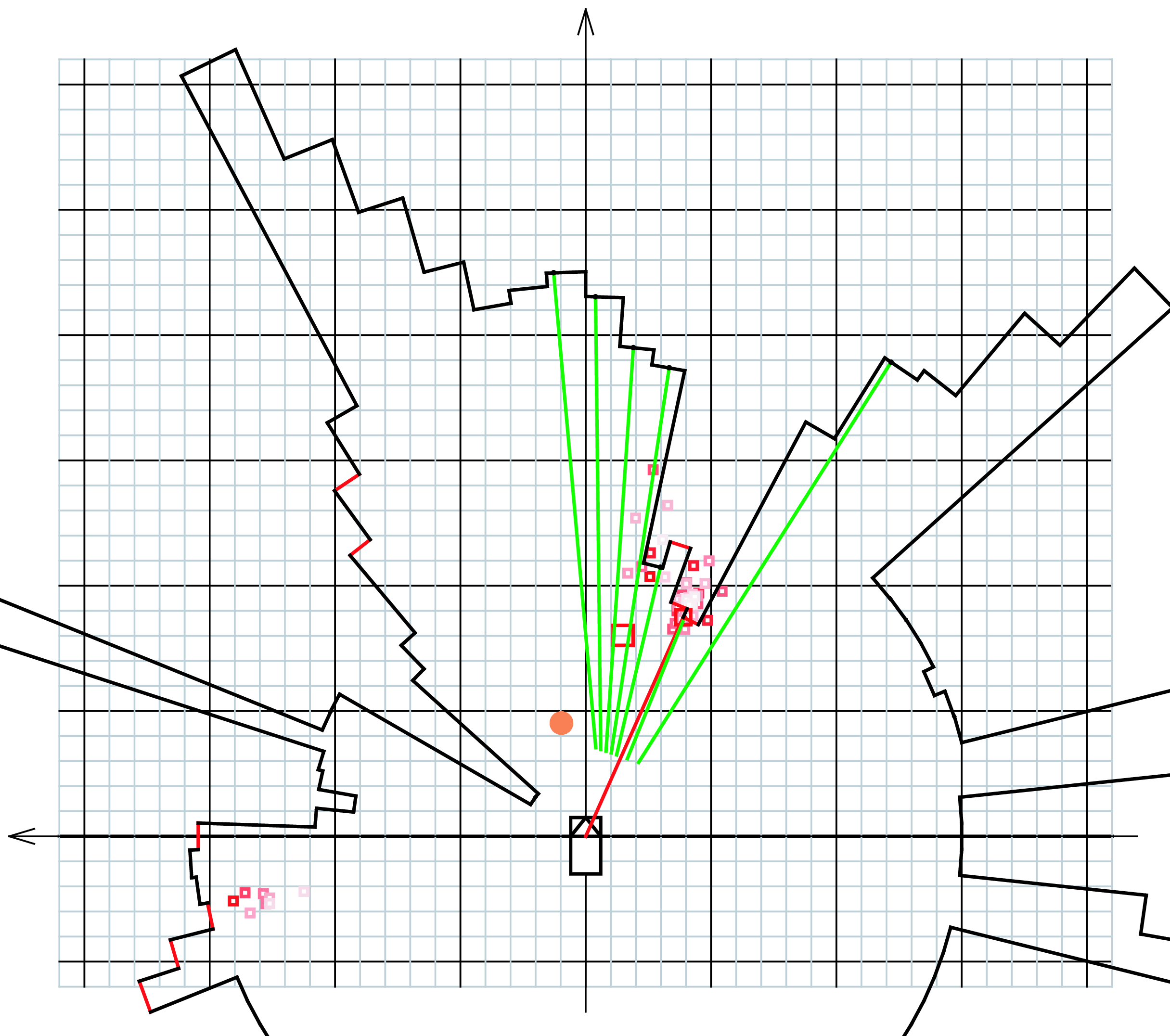} \\
\end{tabular}}
\caption{Top row: The scene from figure \ref{f:scene} seen through the cameras of robots $A$ and $B$. Second row: From each image, the positions of the ball, other agents
    and obstacles are extracted. The images are scanned along lines orthogonal to the horizon for characteristic gradients in the color channels. Bottom row: The agents maintain a continuous analog model of their immediate surroundings by integrating the (noisy) information extracted from the camera images. The graphs show snapshots of this model at the time when the images in a) and b) were taken. } 
\label{f:perception}
\end{figure}

Our experiment uses physical robotic `agents', which roam around freely in an unconstrained in-door environment (see figure \ref{f:scene}). The agents have subsystems for autonomous locomotion and vision-based obstacle avoidance. They maintain a real-time analog model of their immediate surroundings based on visual input (see figure \ref{f:perception}). Using this analog model, the robots track other robots as well as orange balls using standard image processing algorithms. Furthermore the robots have been endowed with a subsystem to segment the flow of data into distinct events and they then build a situation model. There is a short term memory which contains the situation model of the most recent event and a number of past events. 

The robot agents engage in language games - routinised communicative interactions. Two robots walk around randomly. As soon as one sees the ball, it comes to a stop and searches for the other robot, which also looks for the ball and will stop when it sees it. Then the human experimenter pushes the ball with a stick so that it rolls a short distance, for example from the left of one robot to its right. This movement is tracked and analyzed by both robots and each uses the resulting perception as the basis for playing the language game, in which one of the two (henceforth the `speaker') describes the ball-moving event to the other (the `hearer'). To do this, the speaker must first conceptualize the event in terms of a set of categories that distinguishes the latest event from previous ones, for example that the ball rolled away from the speaker and to the right, as opposed to towards the speaker, or, away from the speaker but to the left. The speaker then expresses this conceptualisation using whatever linguistic resources in his inventory cover the conceptualisation best and have been most successful in the past. The game is a success if, according to the hearer, the description given by the speaker not only fits with the scene as perceived by him but is also distinctive with respect to previous scenes.

Agents take turns playing speaker and hearer so that they each gradually develop the competence to speak as well as interpret. No prior language nor prior set of perceptually grounded categories are programmed into the agents. Indeed the purpose of the experiment is to see what kinds of categories and linguistic constructions will emerge, and more specifically, whether they involve perspective marking or not. 

The agents use two additional subsystems to achieve this as described in more detail shortly. The first one performs categorisation and category formation \citep{Harnad:1987}. We use here discrimination trees (as explained further below), although other categorisation methods (e.g. Radial Basis Function networks or Nearest Neighbour Classification) would work equally well. The agents apply categorisation to the sensory channels that directly reflect properties of the visual image computed using standard image processing algorithms, such as start and end-position of the ball, angle of the trajectory, distance traveled by the ball, etc. 
The second subsystem concerns the lexicon. We use a bi-directional associative memory which associates one pattern (here a set of categories) with another pattern (here a word). The associations are weighted with a score because the same pattern may be associated (in either direction) with more than one other pattern. Indeed, one word can have many meanings (synonymy) and several words can be in competition for the same meaning.  In retrieving a target given a source, the association with the highest score is preferred. Neural implementations of bi-directional associative memories have been well studied and shown to be applicable in a wide range of domains \citep{kosko88bidirectional}. 

The behavior of the two subsystems (for categorisation and lexicon lookup) is structurally coupled in that success in the game raises the score both of the categories that were used {\it and} of the lexical conventions that were used to express those categories, so that agents progressively come to share not only their linguistic conventions but also their conceptual repertoires (as extensively shown in \citealp{steels05coordinating}). 

In addition to subsystems for visually perceiving and acting in a dynamically changing world, extracting and memorizing events, discriminating events from previous ones using discrimination trees, and lexicalising these distinctions using a bi-directional associative memory, agents are endowed with a subsystem for egocentric perspective transformation, so that they can reconstruct a scene from the viewpoint of another agent. This requires that they first detect where the other agent is located (according to their own perception of the world) and then perform a geometric transformation of their own world model. Inevitably, an agent's reconstruction of how another agent sees the world will never be completely accurate, and may even be grossly incorrect due to unavoidable misperceptions both of the other robot's position and of the real world itself.  The sensory values obtained by the robots should not be interpreted as exact measures (which would be impossible on physical robots using real world perception) but at best as reasonable estimates. This type of inaccuracies is precisely what a viable communication system must be able to cope with and robotic models are therefore the only way to seriously test and compare strategies and the mechanisms that implement them. 

The following subsections provide some more technical detail and examples of each of these subsystems at work. Readers who are not interested can skip the remainder of this section and immediately look at the results of the experiments on perspective alignment and perspective marking. 

\subsection{Embodiment, Behavior, and Perception}
\label{s:embodiment}

As robots we use the Sony ERS7 AIBO which is a highly complex fully autonomous and fully programmable robot. In addition to the on-board computing power, we use an external computer to control the experiment and engage in some of the symbolic aspects of each robot's behavior. 

\begin{figure}[t]
\centerline{\includegraphics[width=0.7\columnwidth]{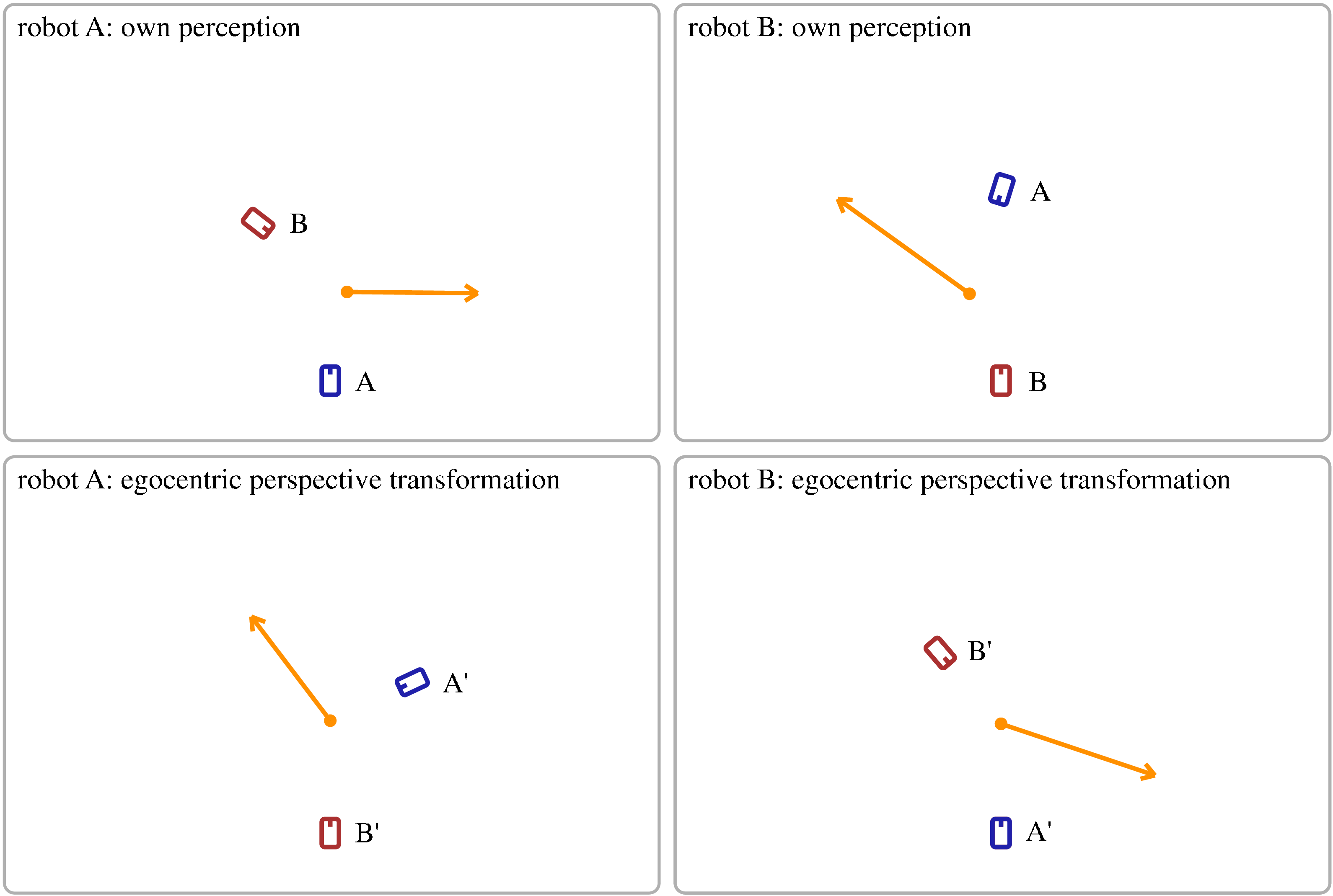}}
\caption{The agents are endowed with the ability to segment the continuous stream of visual data (figure \ref{f:perception}) into discrete event descriptions that make up their situation model. Top row: The event from figure \ref{f:scene} as perceived from robots $A$ and $B$. Bottom row: The result of egocentric perspective transformation. Both robots are able to construct a description of the scene as it would look like from the perceived position of the other robot.}
\label{f:4p}
\end{figure}

Although there has been a lot of progress in robotics during the last years, particularly due to the rise of the `behavior-based approach to robotics' \citep{SteelsBrooks:1994}, doing perception and autonomous behavior with real robots is still an extremely difficult task. We could not have done this experiment without relying on the existing robot soccer software developed by \cite{roefer04germanteam}.  The vision system has to deal with noisy and low resolution (208 $\times$ 160 pixel) images from a robot's camera. Objects like the ball look very different in different places of the environment due to slight differences in illumination. Noisy perception introduces the challenge of maintaining a robust situation model. As the perception can not be always trusted, the resulting position of the ball is only an estimated position gained with probabilistic filtering techniques. As shown in figure \ref{f:4p}, the two robots never perceive the scene in exactly the same way.

Behavior-based control systems \citep{loetzsch06xabsl} were implemented for the physical coordination of the robots. Both robots randomly walk around while avoiding obstacles. Each robot that sees both the ball and the other robot sends an acoustic signal. Robots continue with random exploration until a configuration is reached so that they both see the ball and the other robot and know that the other robot is doing so as well (they establish a joint attentional frame in the sense of \citealp{tomasello95jointattention}). When both robots are ready to observe the scene together, a human experimenter  manually moves the ball. The begin and end point of the trajectory  are recorded and sent to the language system via the wireless network (see figure \ref{f:4p}).

As shown in the bottom row of figure \ref{f:4p}, each robot is able to compute an additional description of the scene from the perspective of the other robot (egocentric perspective transform) so that they are in fact able to compute the situation model from another perspective than their own. Note that this situation model is not always accurate (due to the difficulty of each robot to perceive the perception of the other. In figure \ref{f:4p}
robot $A$'s situation model of $B$ (bottom left in figure \ref{f:4p}) is slightly different from robot $B$'s actual situation model (top right in figure \ref{f:4p}). 

\subsection{Conceptualisation by the Speaker}
\label{s:language-game}

\begin{figure}[t]
\centerline{\begin{tabular}{ll}
a) & b) \\
\includegraphics[width=0.4\columnwidth]{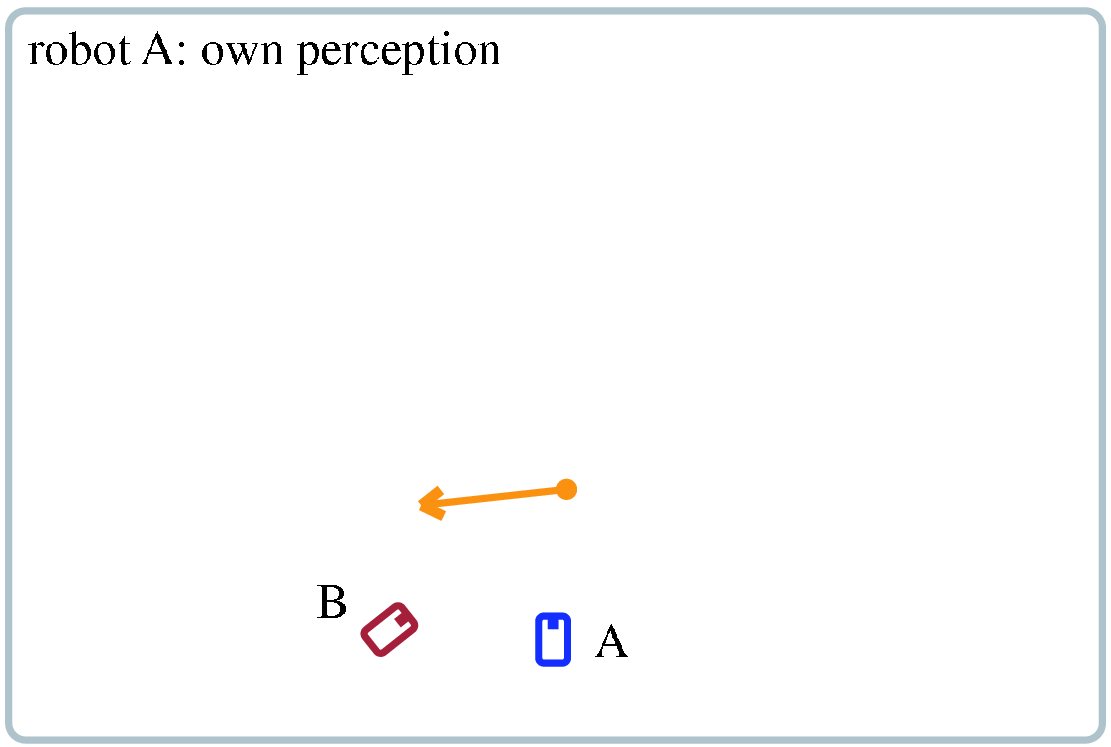} &
\includegraphics[width=0.4\columnwidth]{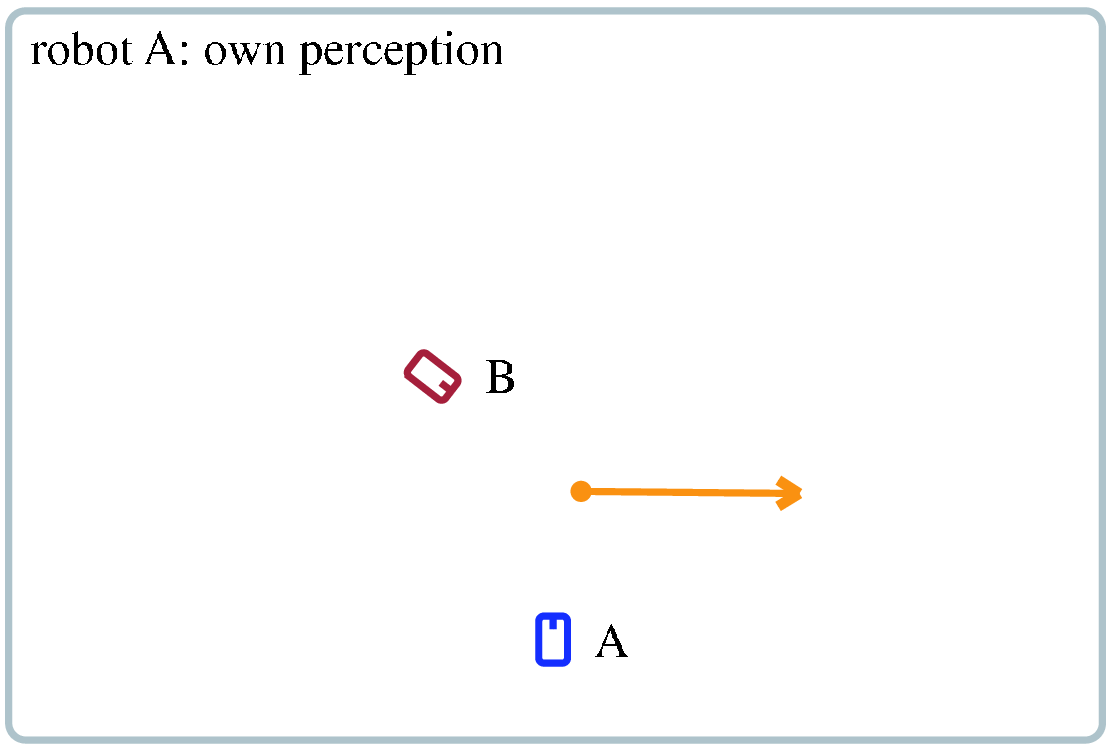}
\end{tabular}}
\caption{Two events as subsequently perceived by robot $A$. The goal of conceptualization is to find a set of categories that discriminate the recent event (b) from the previous event (a).}
\label{f:conceptualization}
\end{figure}

The goal of the conceptualisation subsystem is to come up with the {\it meaning} to be expressed by the speaker. This meaning should be such that it discriminates the topic (the most recent event) from the other events in the context. Conceptualisation decomposes into three subsystems. The first one extracts a battery of features from the perceived scene. The second categorises the objects in the context based on these features, and the third subsystem finds out which categories are discriminative. Because the speaker can compute the scene from the perspective of the hearer, he will not only conceptualise from his own perspective but also from that of the hearer so that he can determine whether perspective needs to be marked or whether he is going to be more successful to describe the scene from the perspective of the hearer because that is more salient and can be done with more established categories. 

It is helpful to see the operation of the different subsystems for a concrete example. 
We take the 4116th interaction from a series in a population of 5 agents. Agents 3 and 4 were randomly drawn from the population, agent 3 was randomly assigned to be the speaker and ``used'' robot body A. Agent 4 was the hearer (robot B). Both have perceived two events (for robot $A$ shown in fig. \ref{f:conceptualization}).

\begin{figure}[t]
\centerline{\includegraphics[width=0.7\columnwidth]{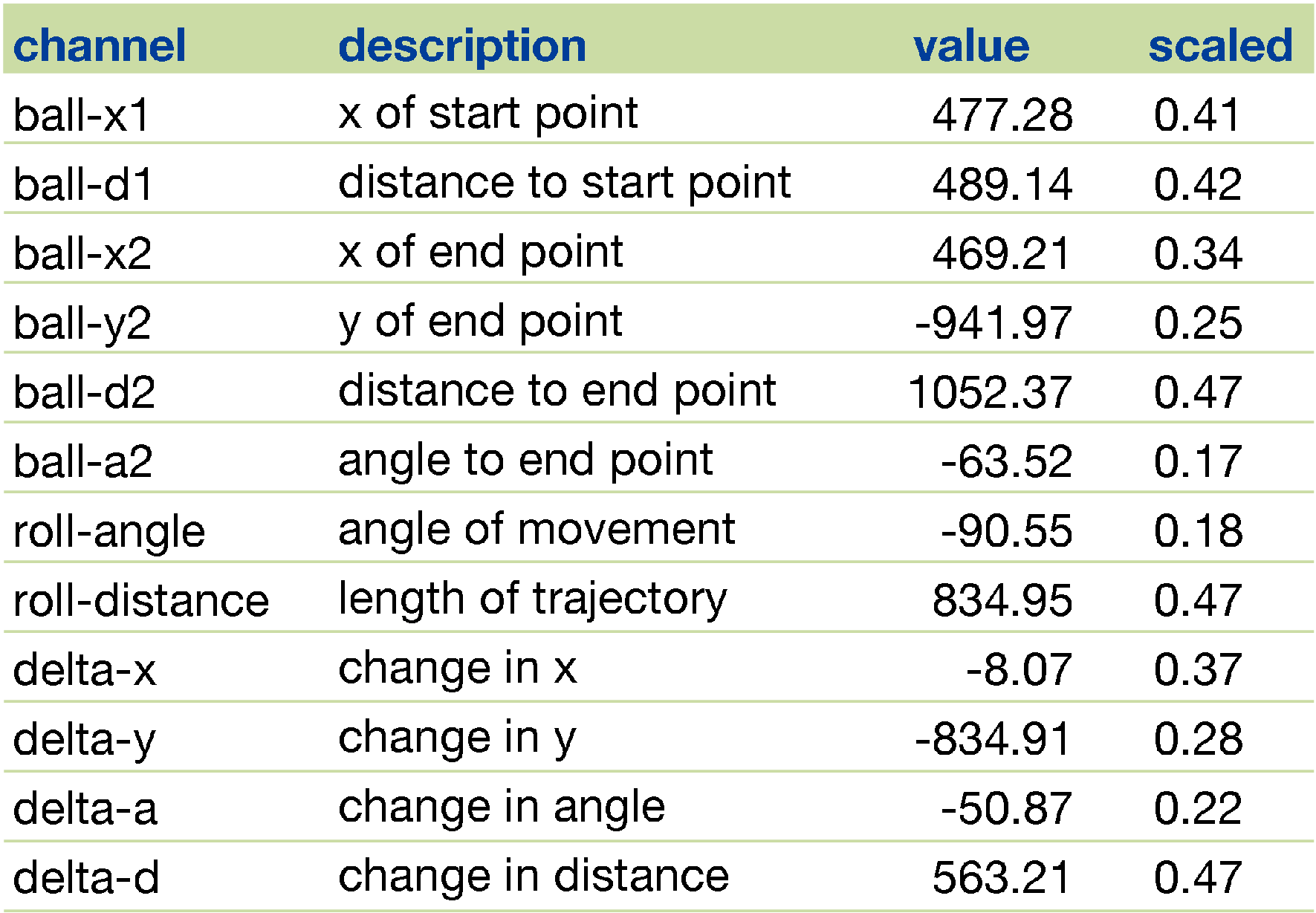}}
\caption{The feature values for the event in figure \ref{f:conceptualization}a).}
\label{f:channels}
\end{figure}

Categorisation operates over 12 feature channels which are calculated for each event based on straightforward signal processing and pattern recognition algorithms (see figure \ref{f:channels}). For example, channel \tt ball-x1 \rm is the x component of the start position of the ball, \tt ball-y2 \rm is the y position at the end of the movement, \tt delta-a \rm is the change in angle of the ball, and so on. For ease in further processing and in order to be able to compare features, each channel value is scaled within the interval [0...1]. 1 means that it is a very high channel value (with the respect to the observed distribution for that particular channel) and 0 a very low value.

%\subsubsection{Categorisation by the Speaker}

Categorisation itself is performed with a discrimination tree approach described in more detail in \citep{steels96perceptually}.  In order to help the hearer guess what the speaker meant, the most  \it salient \rm feature is chosen. Saliency is computed as the minimum distance of the feature values of the topic to the average feature values of other events in the context:\\

\begin{tabular}{lccclccc} 
  channel & \tt ball-y2 \rm & \tt delta-y \rm & \tt roll-angle \rm & \dots & \tt ball-x1\rm & \tt ball-d1 \rm \\
  saliency & 0.72 & 0.70 & 0.52 & \dots  & 0.00 & 0.00 
\end{tabular}\\

As easily seen in figure \ref{f:conceptualization}, the features \tt ball-y2 \rm (end position left/ right) and \tt delta-y \rm (change towards left/ right) are much more salient than \tt ball-x1 \rm (start position far/ close) \rm and \tt ball-d1 \rm (distance to the ball at the beginning).   

\begin{figure}[t]
\centerline{\includegraphics[width=0.9\columnwidth]{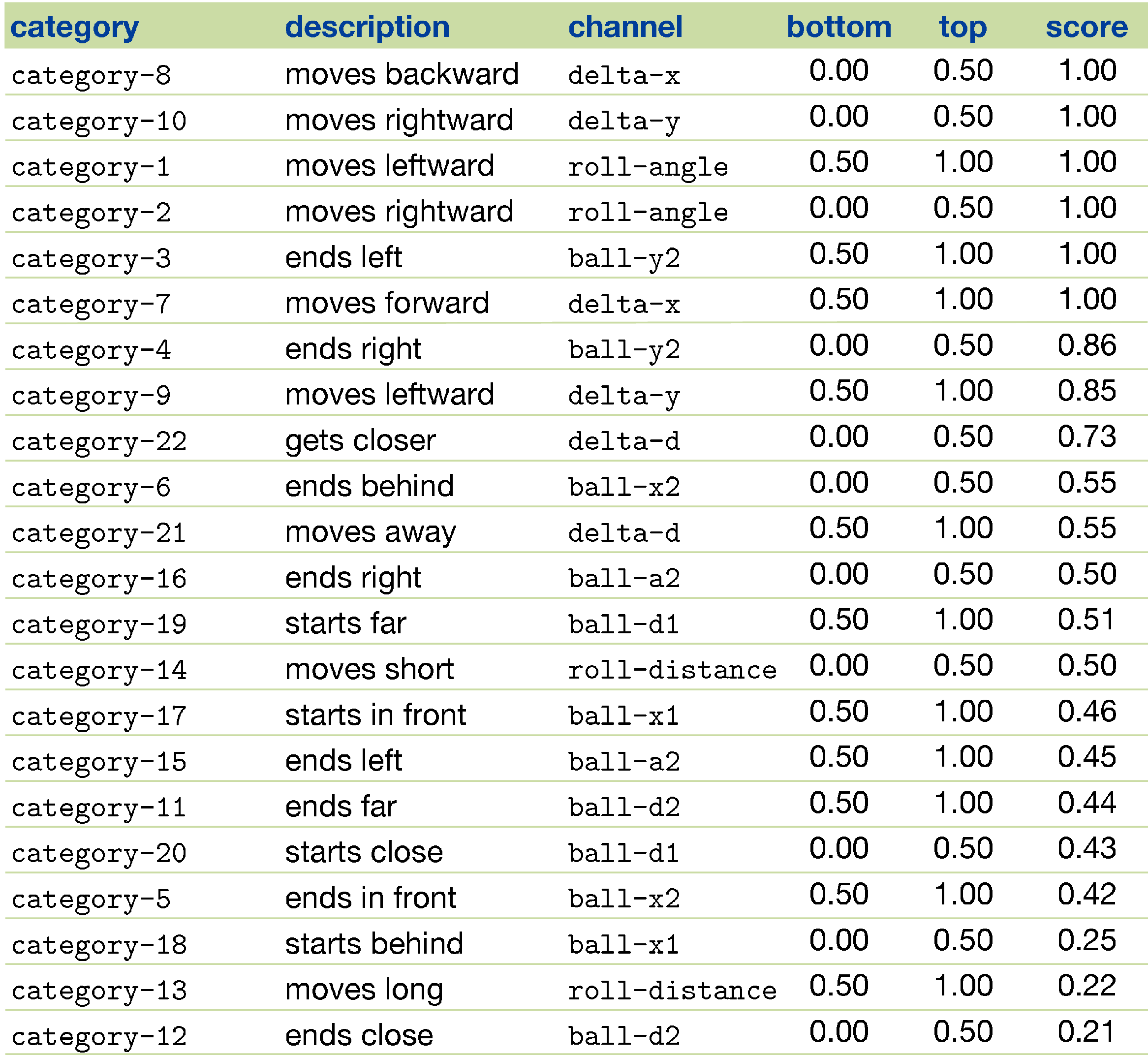}}
\caption{The ontology of agent 3 after 4412 games.}
\label{f:ontology}
\end{figure}

There is a discrimination tree for every feature channel. Each tree divides the range of possible values into equally sized regions, and every region carves out a single category. For example for agent 3 the category \tt category-4 \rm covers the interval \tt [0,0.5] \rm on feature channel \tt ball-y2 \rm (figure \ref{f:ontology}). The set of all categories of an agent is called his ontology. Every category in the ontology has a score which is based on past success in the language games. Through adjustements of the score, agents progressively become aligned  because the score also reflects not only the categories that are relevant in the scenes that they encounter but also those that are commonly used in the group.  

%\subsubsection{Discrimination}
In order to find a discriminating category, the categories for the most salient feature(s) are computed and then those categories retained that are unique for the topic. In the present  example, this is \it the ball ends right \rm (\tt category-4\rm). When there is no discriminating category for the most salient feature channels in the ontology, the ontology is extended by refining a category applicable to the topic. Refinement of a category $c$ happens by dividing the region of $c$ into two equally sized subregions, which then yield two new subcategories. In the current experiment, the tree depth of the ontology never had to go deeper than one however. 

We use predicate-calculus notation (in prefix) to display the `meaning' that is being expressed by the speaker (and reconstructed by the hearer). The predicates consist of all the categories in the ontology of the agent and the arguments are the event and the truth value. Here is an example: 
\begin{verbatim}
(category-4 event-16462 t)
\end{verbatim}

\subsection{Perspective Reversal by the Speaker}

In some of the experiments we investigate the role of perspective alignment and perspective reversal. As mentioned earlier, we have endowed the agents with the capacity of egocentric perspective transform, so that they can not only build up a situation model of themselves but also of what the other robot is supposed to see. If that is the case, the speaker can check whether the discriminating category of the scene which is valid for his own situation model also holds for that of the hearer. If so, perspective does not need to be marked (the perception of that feature of the scene is shared). Otherwise, the meaning to be expressed is extended with an additional predicate (`\tt own-perspective\rm') to specify that the perspective is seen from the one of the speaker.  In the example, the category is not discriminative for the situation model from the hearer's perspective, in fact it does not even hold in this model (the ball moves to the left in both events for the hearer). Hence the meaning is expanded by a perspective indicator:

\begin{verbatim}
(category-4 event-16462 t)
(own-perspective event-16462 t) 
\end{verbatim}

Alternatively, the speaker can completely conceptualise the scene from the viewpoint of the hearer and will choose it if it can be done with a more salient feature channel and based on a more established category. As one can see in figure \ref{f:4p} (left bottom), for the assumed perspective of the hearer (robot $B$) the change in x position (channel \tt delta-x \rm) is the most salient channel, and the appropriate category (which happens to be \tt category-7 \rm or \it moves forward\rm) can now be used.

Meanings are ranked based on saliency and category score. 
The description with the highest score is then used in lexicalization. For the present case we have: 

\begin{verbatim}
(category-4 event-16462 t)   0.393  ; from own perspective 
(category-7 event-16462 t)   0.363  ; from other perspective 
\end{verbatim}
So the first meaning is the best one from the viewpoint of conceptualisation. 

In the third experiment to be discussed later, the perspective is explicitly marked, which implies that it must be part of the meaning transmitted from the conceptualisation subsystem to the lexical subsystem. Perspective is represented with two predicates own-perspective 
and other-perspective, as in: 
\begin{verbatim}
(category-7 event-16462 t)
(other-perspective event-16462 t)
\end{verbatim}

\subsection{The Lexicon for the Speaker}

\begin{figure}
\centerline{\includegraphics[width=0.5\columnwidth]{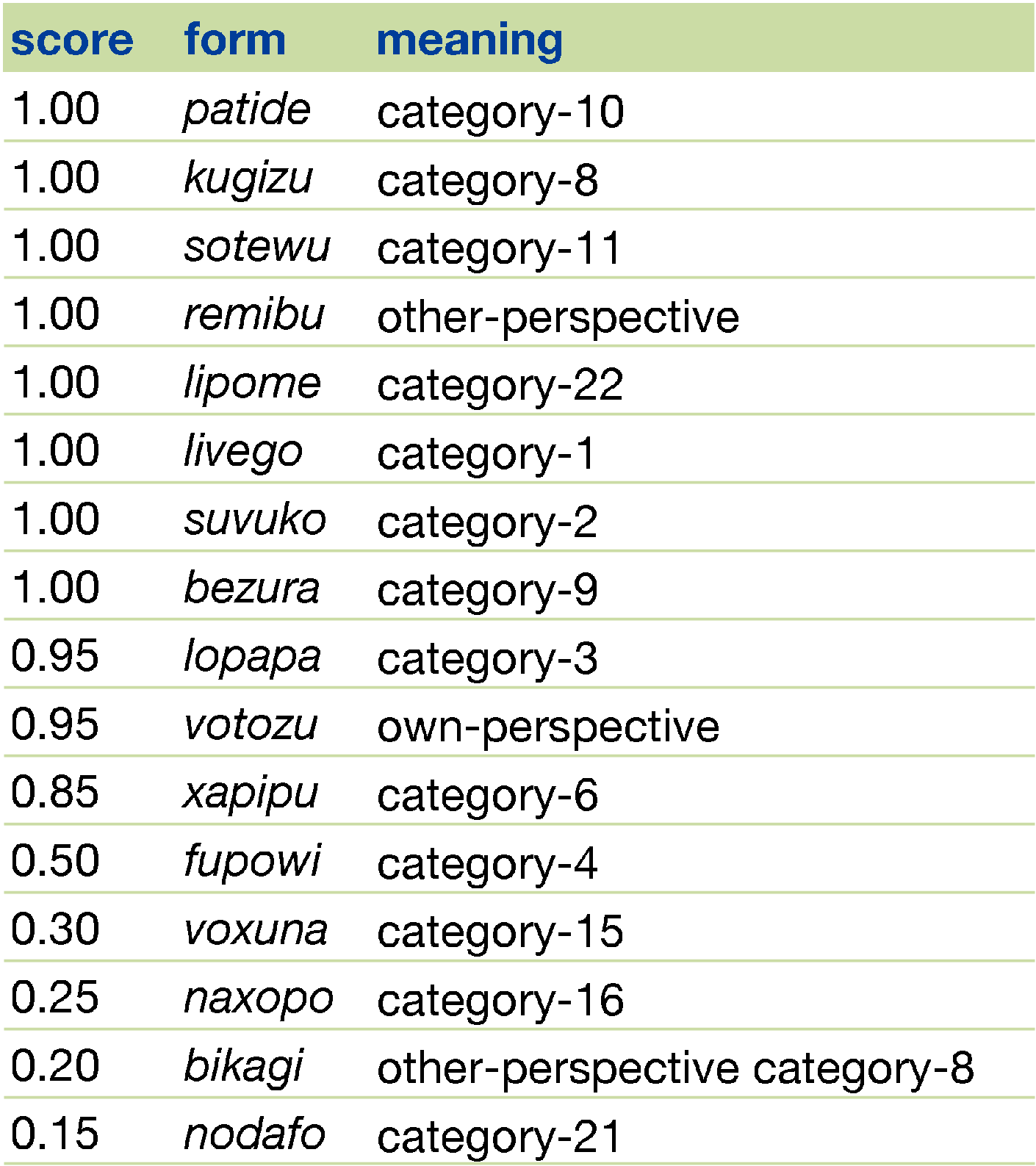}}
\caption{The lexicon of agent 3 after 4412 games.}
\label{f:lexicon}
\end{figure}

Each agent has a linguistic inventory, the lexicon (figure \ref{f:lexicon}). It is a bidirectional associative memory that associates abstract meanings (predicates and arguments with variables) to forms (words). Each association has a weight which acts as a score, reflecting how well the word involved had success in previous language games. We know from many earlier experiments that a reinforcement learning approach using lateral inhibition is an effective way to self-organise a lexicon \citep{steels01ieee}. The speaker selects the smallest set of words that covers the complete meaning to be expressed (in the present example this is \it fupowi votozu\rm). In case there are alternative solutions, the form-meaning pairs with the highest score are used. Whenever the speaker does not have a word for the whole meaning or part of it, a new word is invented by combining random syllables and associating them with the uncovered meaning. 

\subsection{Lexicon Lookup and Conceptualisation by the Hearer}

The hearer uses the same knowledge sources (lexicon and ontology) but in the reverse direction. He looks up the words in the lexicon and reconstructs the possible meanings. Usually there are several possibilities as words may be ambiguous. Next he applies to interpret these meanings by matching them against the (reconstructed) situation model of the speaker and then his own situation model. 

\begin{figure}[p]
\centerline{\rotatebox{90}{\includegraphics[width=1.03\textheight]{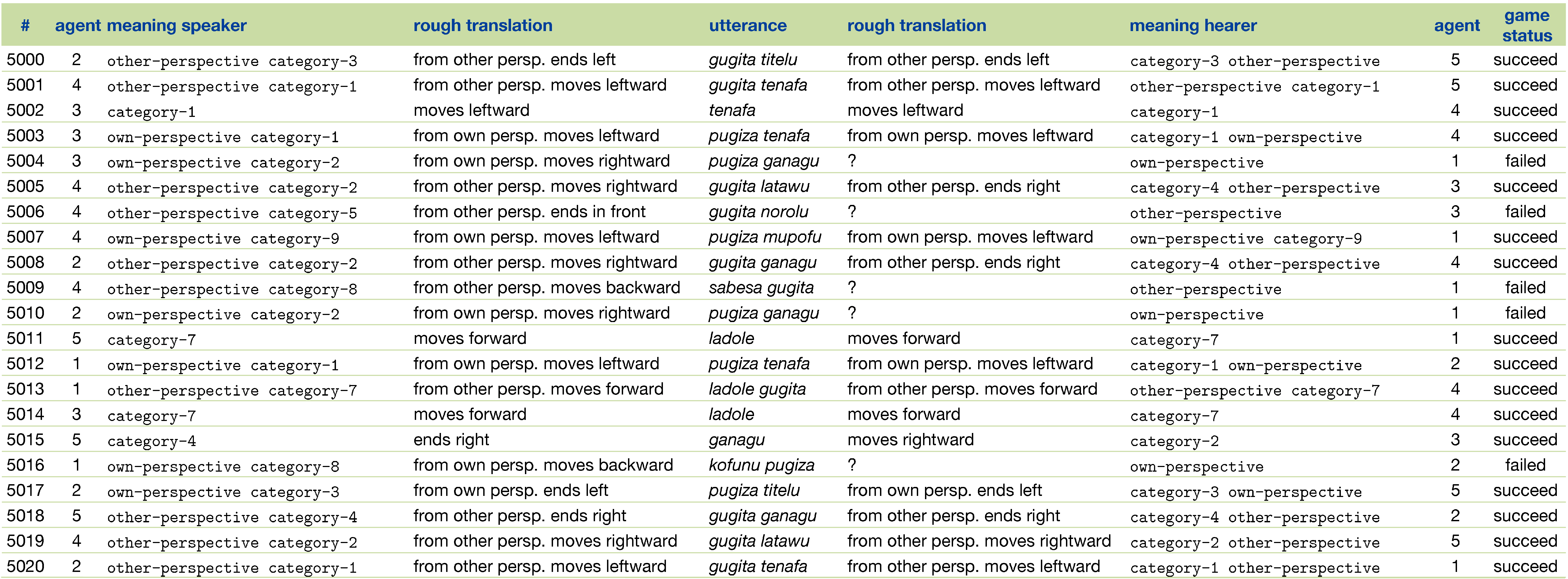}}}
\caption{Subsequent interactions in a population of 5 agents (games 5000--5020).
\label{f:subsequent}}
\end{figure}

\subsection{Feedback}

A game is a success if the hearer knows all the words in the utterance and if the
extracted meanings are true and discriminating for the current event. Everything else is a failure. Communicative success is the only measure that drives the coherence of perceptual categories and lexical items among the agents of a population. Therefore, each category and meaning-form association has a score that reflects its overall success in communication.

After a successful game, the score of the lexical entries that were used for production 
or parsing is increased by 0.05. At the same time, the scores
of competing lexical entries with the same form but different meanings are decreased
by 0.05 (lateral inhibition). In case of a failure, the score of the involved items is 
decreased by 0.05. This scoring adjustement not only acts as a reinforcement learning mechanism but also as priming mechanism so that agents gradually align their lexicons in consecutive games. 

When the hearer does not know one of the words of the utterance, he conceptualizes the scene himself by using the meanings that are already known from the utterance and the additional meanings are then associated with the unknown word(s). This step leads to a kind of replicator dynamics, because words invented or used by the speaker become part of the repertoire of the hearer which could then use it in subsequent interactions.  

Agents not only play a single game, but take turns playing games (see figure \ref{f:subsequent}) and it is through these consecutive games that a consensus gradually arises in the group. Not only the lexicons become aligned but also the ontologies. More and more agents will prefer to use the same conceptualisation in the same sort of circumstances and use similar words for similar meanings.  

\section{Experimental Results for Perspective Alignment}

As stated in the introduction, we want to show why perspective is relevant in spatial language and how agents manage to align and mark perspective.    

\subsection{The Need to Consider Perspective} 

We begin with a first experiment to argue the first point stated in section 1: 
{\it As soon as agents are embodied, they necessarily have a specific view on the world and spatial language becomes impossible without considering perspective.} It is straightforward to do a very clear experiment with the mechanisms introduced in the previous section. 

\begin{figure}[p]
\centerline{\includegraphics[width=0.8\textwidth]{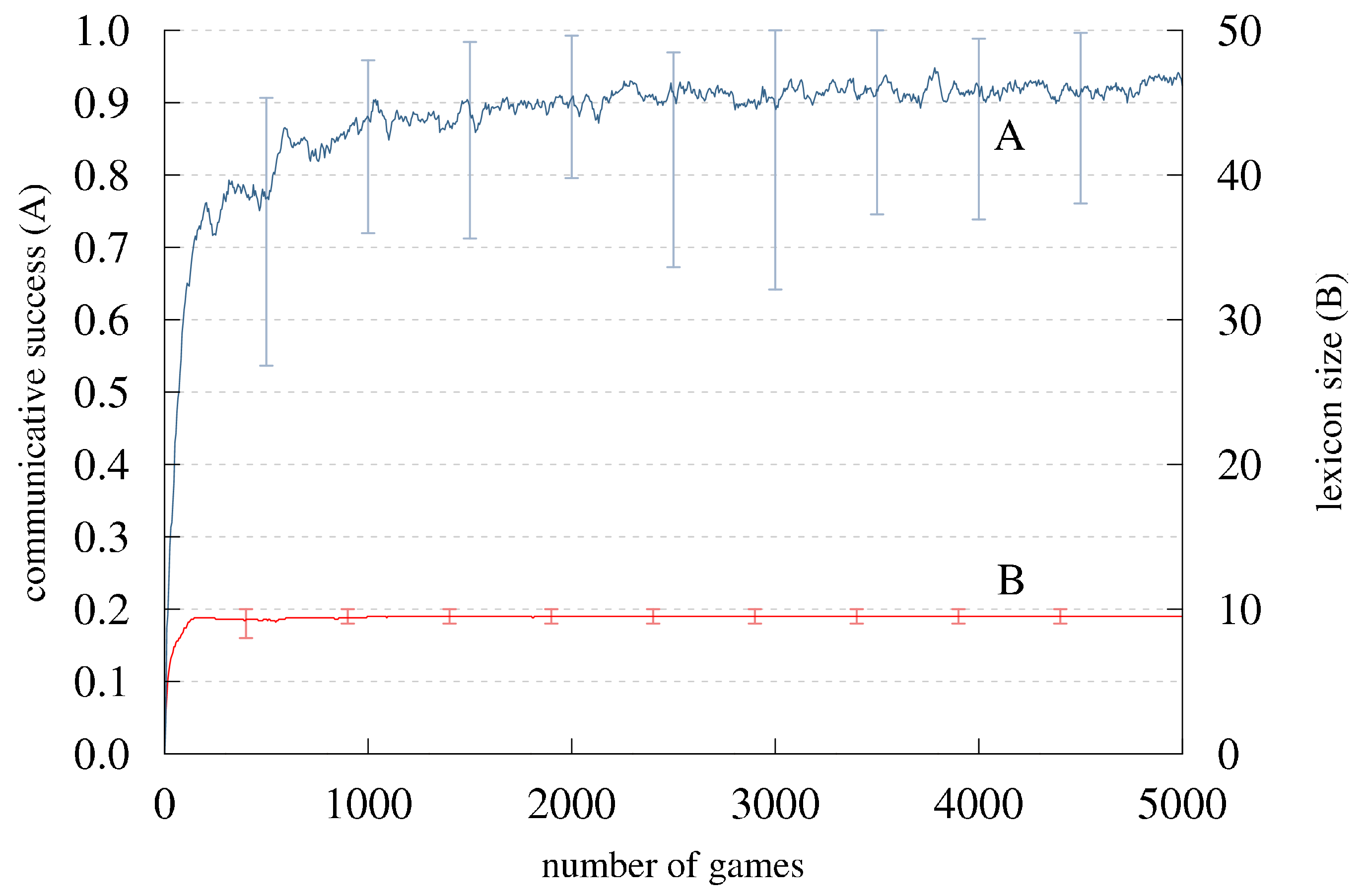}}
\caption{Agents have the same sensory information and hence share their situation model. They quickly self-organise a lexicon and ontology.}
\label{f:same-perception-result-no-reversal}
\end{figure}

First we show in a baseline condition that the cognitive mechanisms proposed earlier for behavior, perception, conceptualisation, and lexicalisation are adequate when both agents engaged in a dialogue perceive the scene through the same camera and hence have exactly the same situation model. Although there is still some form of embodiment here (in the sense of using real vision and real world action), it is not `real' embodiment in the sense of each agent having their own body. As shown in figure \ref{f:same-perception-result-no-reversal}, communicative success quickly increases to 90\% and the average lexicon size of the agents is 10. These results are based on 10 runs of 5000 language games each. We show the average and the variance. So this experiment shows convincingly that the mechanisms proposed here work properly.  

\begin{figure}[p]
\centerline{\includegraphics[width=0.8\textwidth]{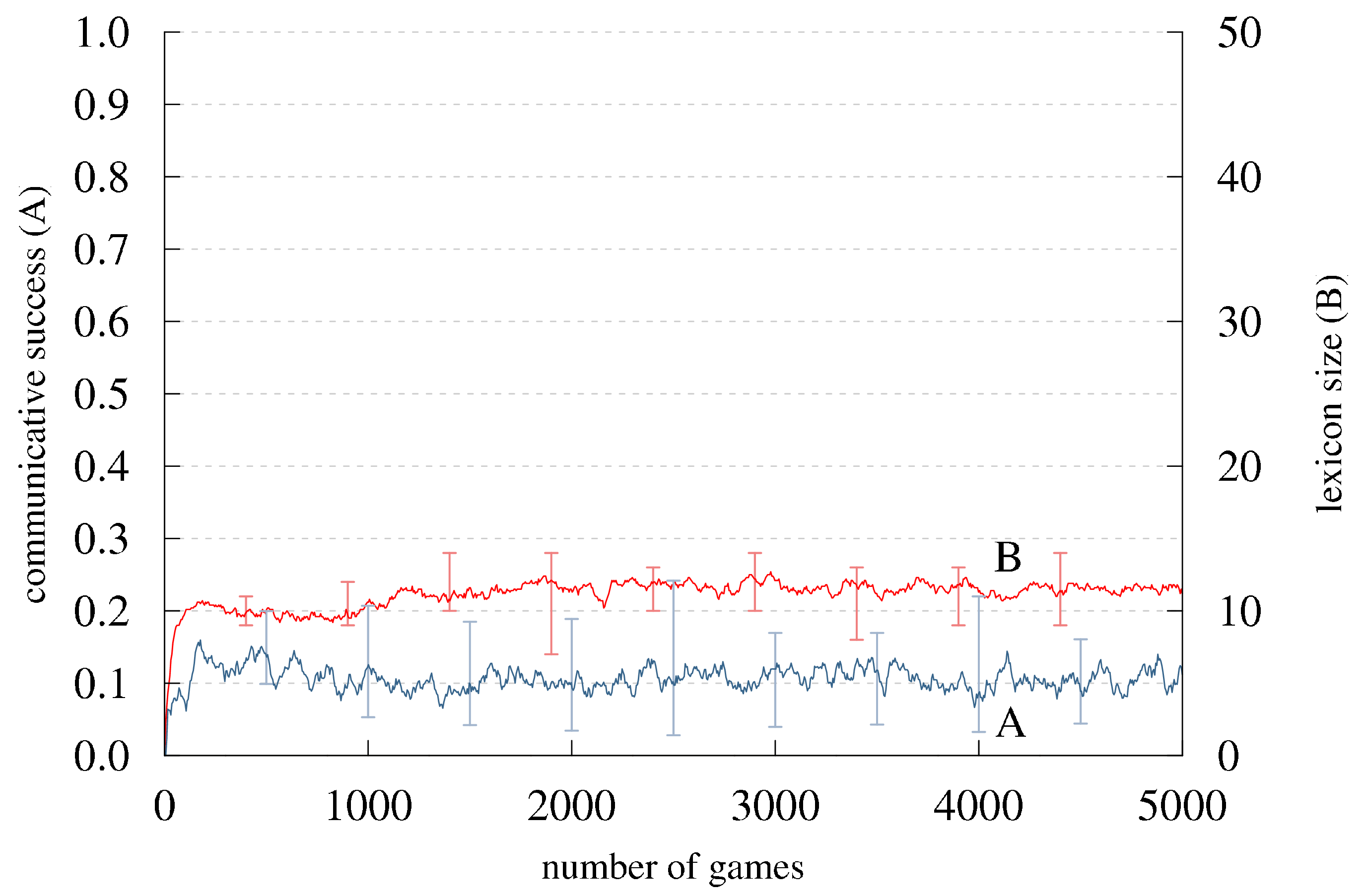}}
\caption{Agents do not share sensory stimuli and do not consider perspective. The system does not come off the ground.}
\label{f:result-no-reversal}
\end{figure}

In the next condition, the agents perceive the scene through their own camera but they do not take perspective into account. The results are shown in figure \ref{f:result-no-reversal}. Now they do not manage to agree on a shared set of spatial terms. Communicative success does not reach 10\%. This clearly proves the first thesis, namely that grounded spatial language without perspective does not lead to the bootstrapping of a successful communication system for this kind of communicative task.  
 
\subsection{Perspective without marking} 

The next argument we wanted to make is the following: {\it Perspective alignment is possible when the agents are endowed with two abilities: (i) to see where the other one is located, and (ii) to perform a geometric transformation known as Egocentric Perspective Transform.} Both of these abilities have been implemented for the robots as explained earlier and so it is now possible to do an experiment that exercises these mechanisms. 

When agents are able to perform egocentric perspective transformation and when the allocentric situation model is used as well in conceptualization, a successful communication system indeed emerges (figure \ref{f:result-no-marking}.) Communicative success again reaches 90\% and the lexicon stabilizes. This is even without marking perspective. The reason the agents are nevertheless successful is because they continuously check from each perspective what a possible meaning or a possible interpretation might be. So we have at once an answer to the question how it is possible for two partners in dialogue to align perspective even if there is no explicit marking.

\subsection{The Role of Perspective Marking}

\begin{figure}[p]
\centerline{\includegraphics[width=0.8\textwidth]{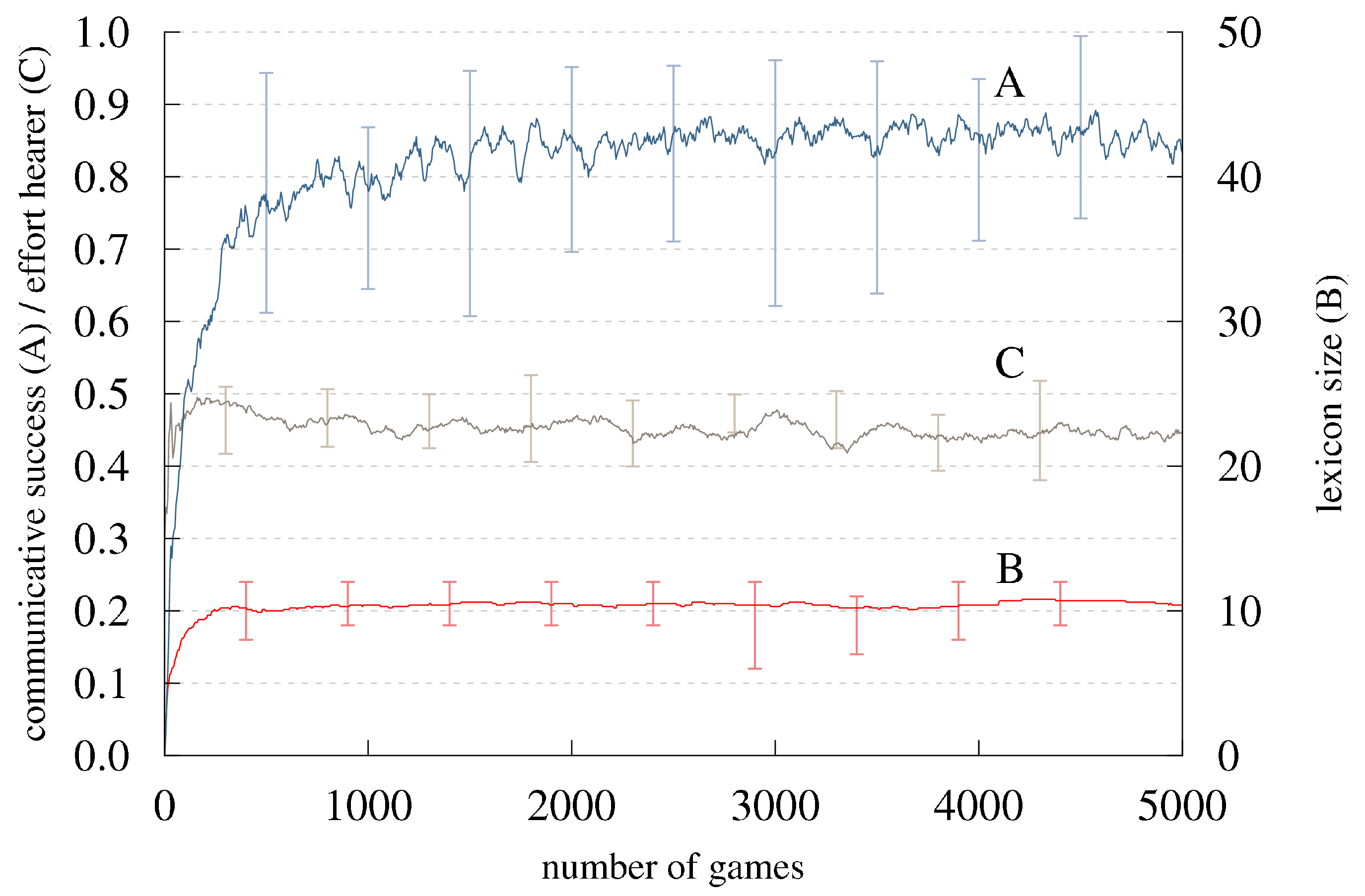}}
\caption{Agents are able to adopt the interlocutor perspective but do not mark their perspective choice in language. They manage again to self-organise a spatial language system.}
\label{f:result-no-marking}
\end{figure}

We now perform a third experiment to examine the third thesis: {\it Perspective alignment takes less cognitive effort if perspective is marked.} In the previous experiment, the hearer has to guess (by trying to interpret the utterance for both perspectives) which perspective was used
and the speaker has to compute both perspectives to make sure he chooses the one that will have most success with the hearer.  This obviously results in a higher cognitive effort for the hearer. Cognitive effort is defined as the average number of additional perspective transformations that the hearer has to perform and was shown already in figure \ref{f:result-no-marking}. 
 
Now we change slightly the language architecture for each agent. 
The chosen perspective is made explicitly a part of the meaning so that it becomes lexicalised.
For example, as in: 

\begin{verbatim}
(category-7 event-16462 t)
(other-perspective event-16462 t)
\end{verbatim}

and this will automatically lead to an expression of perspective. Note that 
the lexicon formation process is completely general. It tries to cover the complete meaning with the smallest number of words  and invents new words for parts that are not yet covered. Nevertheless we see that separate words emerge for perspective in addition to words where perspective is part of the lexicalisation of the predicate. This is similar to natural language where in ``the ball to my left", ``my" is a general indicator of perspective, whereas in the German ``hinein" (``into'' from outside perspective) versus ``herein" (``into'' from inside perspective) or English ``come" and ``go", perspective is integrated in the individual word. So this experiment explains why perspective marking occurs in human languages and why sometimes we find specific words for it. 

\begin{figure}[p]
\centerline{\includegraphics[width=0.8\textwidth]{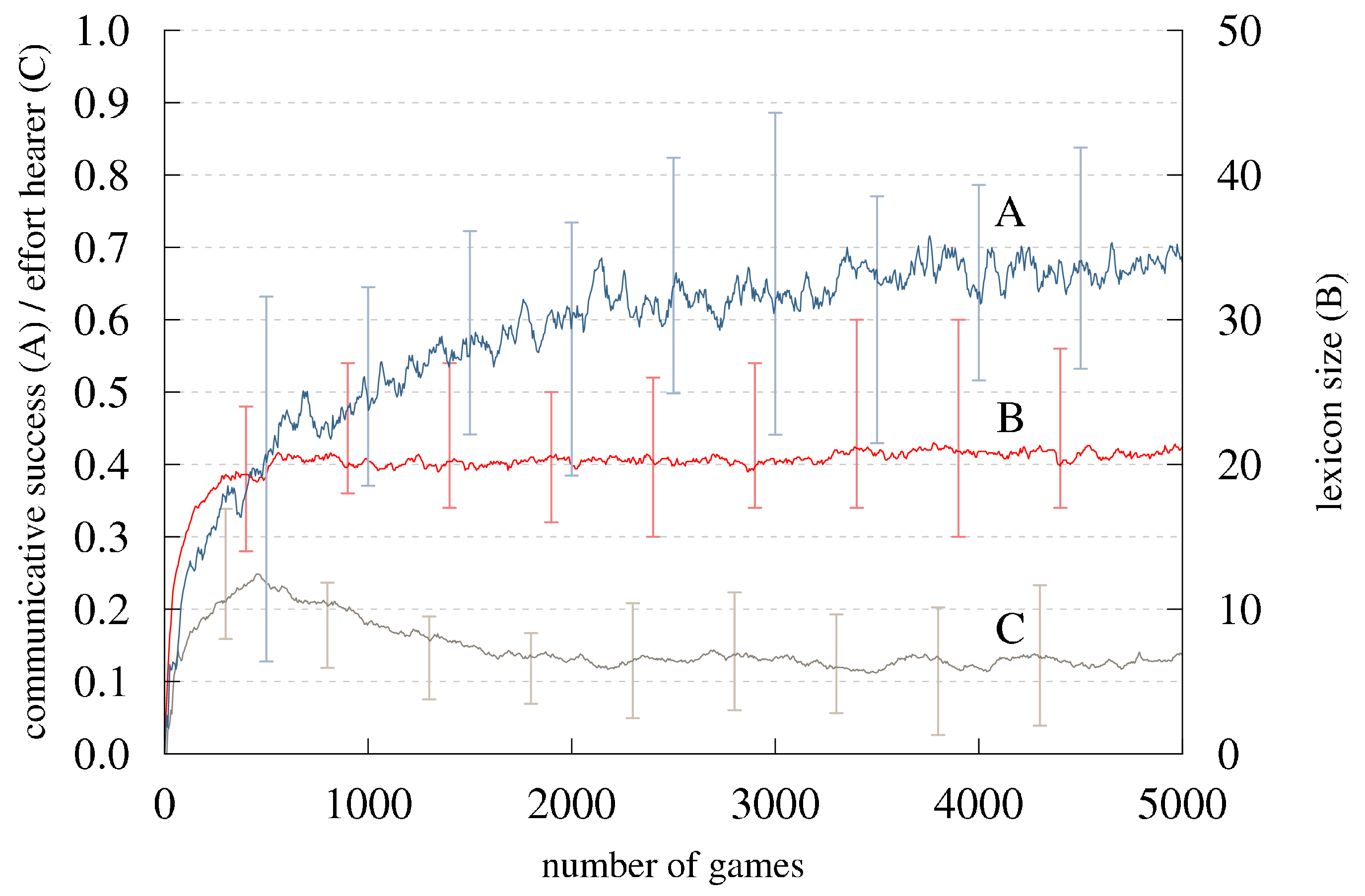}}
\caption{Agents now additionally mark their perspective choice in language. This maintains communicative success but results in a decrease of cognitive effort.}
\label{f:result-marking}
\end{figure}

As shown in figure \ref{f:result-marking}, communicative success remains high but the cognitive effort dramatically decreases compared to the earlier experiment. Communicative success is not as high as in the previous experiment without perspective marking (figure \ref{f:result-no-marking}). This is due to the fact that the learning problem is harder for the agents as they additionally have to agree on a set of perspective markers or words that incorporate domain categories and a perspective marker, but if we look at a longer series of games we see that a similar level of success is reached. We have moreover a more compact lexicon as  in the previous experiment. 

\section{Conclusion}

This paper is significant from two points of view. On the one hand, it shows a novel way to investigate spatial language and perspective, namely by doing experiments in which physically embodied agents (robots) are endowed with a `language faculty' that allows them to bootstrap a communication system autonomously (i.e. without human intervention) and from scratch. This rather new methodology is complementary to empirical observations of human dialogue and helps us to develop and test `mechanistic' theories of dialogue \citep{cangelosi02computer}. On the other hand, we could show very precisely why perspective is essential for spatial language, how speaker and hearer could align perspective - even without marking-, and why and how perspective could become explicitly marked as part of spatial dialogue. 

\paragraph{Acknowledgements}. The experiments rest on the Fluid Construction Grammar framework \citep{steels06unify}, which is highly complex software for language processing to which Nicolas Neubauer and Joachim De Beule have made major contributions. The authors also thank the members of the ``GermanTeam'' for providing their robot soccer software and Remi van Trijp for editorial help with the paper. This research was funded and carried out at the Sony Computer Science Laboratory in
Paris with additional funding from the EU FET ECAgents Project IST-1940.

\bibliographystyle{abbrvnat}
\bibliography{../references}

\end{document}